\documentclass[letterpaper]{article}
\usepackage{aaai}
\usepackage{times}
\usepackage{helvet}
\usepackage{courier}
\usepackage{microtype}
\usepackage{latexsym}
\usepackage{graphicx}
\usepackage{fullpage}
\usepackage{color}
\usepackage{multicol} 
\usepackage{url}
\usepackage[pdftex]{hyperref}
\usepackage{wasysym}
\usepackage[hang,flushmargin]{footmisc}
\usepackage{enumitem}
\begin{document}
%
\title{Challenging On Car Racing Problem from OpenAI gym}
\author{Changmao Li\\
Emory University\\
\textit{changmao.li@emory.edu}\\
}
\maketitle
\begin{abstract}
\begin{quote}
This project challenges the car racing problem from OpenAI gym environment. The problem is very challenging since it requires computer to finish the continuous control task by learning from pixels. To tackle this challenging problem, we explored two approaches including evolutionary algorithm based genetic multi-layer perceptron and double deep Q-learning network. The result shows that the genetic multi-layer perceptron can converge fast but when training many episodes, double deep Q-learning can get better score. We analyze the result and draw a conclusion that for limited hardware resources, using genetic multi-layer perceptron sometimes can be more efficient.
\end{quote}
\end{abstract}
\section{Project Introduction}

This project is from CarRacing-v0 challenge in OpenAI gym environment. CarRacing is one of the continuous control task to learn from pixels. State consists of 96x96 pixels. Reward is -0.1 every frame and +1000\/N for every track tile visited, where N is the total number of tiles in track. For example, if you have finished in 732 frames, your reward is 1000 - 0.1*732 = 926.8 points. Episode finishes when all tiles are visited. Some indicators shown at the bottom of the window and the state RGB buffer. From left to right: true speed, four ABS sensors, steering wheel position, gyroscope. It has four controls:up arrow to acceleration, left arrow to steer left, right arrow to steer right and down arrow to brake. 

Several methods has been tried for this project. On the leaderboard of OpenAI gym, there are two people submitted their results considered to solve the problem(this problem defines "solving" as getting average reward of 900 over 100 consecutive trials.). One of them uses the simple feed forward network and the other uses deep Q-learning network. They all claim they solved the problem. For exploring more possible methods, we introduce evolutionary algorithm based genetic multi-layer perceptron and double deep Q-learning network to tackle this problem.

\section{Methodology}

We explored two approaches including evolutionary algorithm based genetic multi-layer perceptron and double deep Q-learning network. For evolutionary algorithm based genetic multi-layer perceptron, we implement it from the scratch; For double deep Q-learning network, we implement it based on demonstration from original paper \cite{HasseltGS15} and the code from Github which implements the basic DDQN to make sure that our implementation for this project is right.

\subsection{Evolutionary Algorithm Based Genetic Multi-Layer Perceptron}

This implementation is from scratch but part of code based on the basic implementation of car racing code since we need to acquire the result parameters after playing. We use evolutionary algorithm based genetic multi-layer perceptron \cite{Maaranen:2004} to implement the computer agent and see the result. 

\subsubsection{Model input and output}
The model receives the following inputs: car speed, car angle, wheel angle, speed direction, car angular velocity, the curvature of the road at n sample points ahead of the car and distance between the car and the center of the road. The model produces the following outputs: acceleration, steer left, steer right, brake. We use genetic method to optimize the neural network which is to encode weights as DNA encodin                                                            
g and update weights like genetic mutation and cross over. 

\subsubsection{Genetic Multi-Layer Perceptron(MLP)}

\begin{figure}[ht!]
\centering
\includegraphics[width=\linewidth]{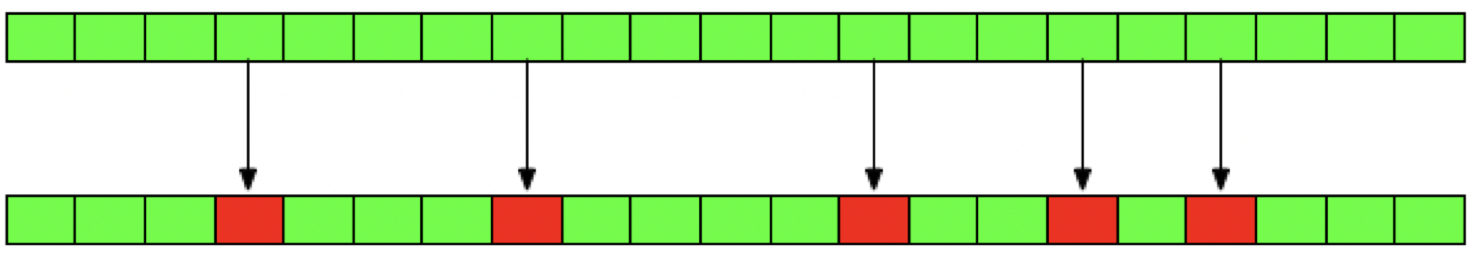}
\caption{DNA mutation}
\label{fig:mutation}
\end{figure}

\begin{figure}[ht!]
\centering
\includegraphics[width=\linewidth]{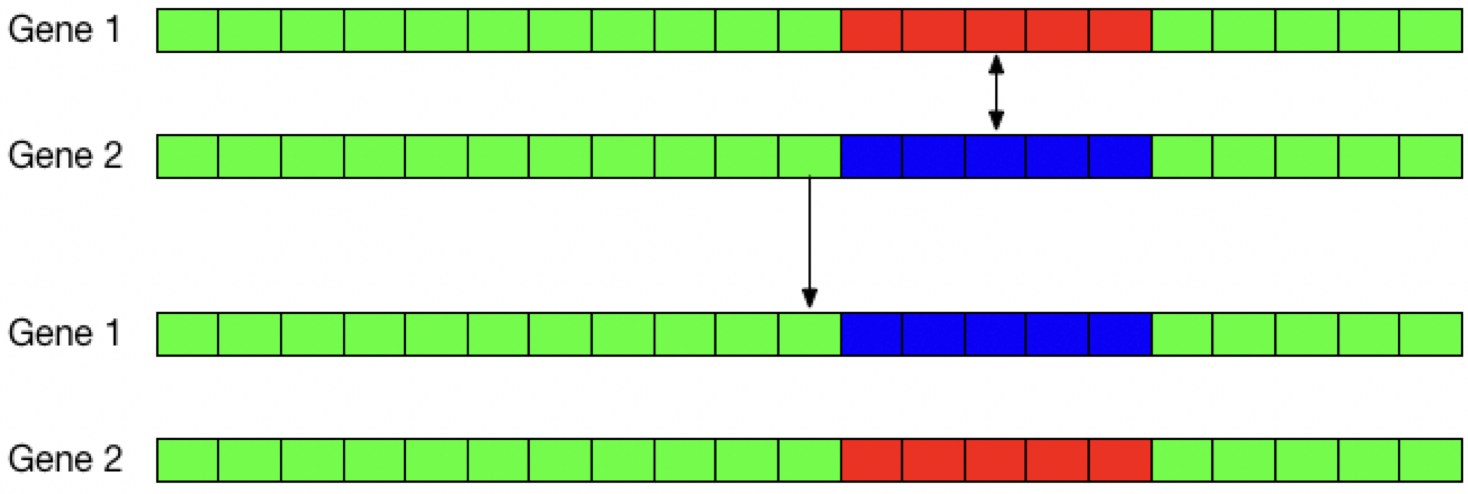}
\caption{DNA crossover}
\label{fig:crossover}
\end{figure}

The genetic MLP is implemented from scratch with two methods to update weights: gene mutations and gene crossover.  Figure \ref{fig:mutation} shows the mutation process, which is to change weights on random encoding position based on the value generated by normal distribution. Figure \ref{fig:crossover} shows the crossover process, which is randomly to replace random segment of random two parent weights encoding.

We implement this self-defined feed-forward network as our neural network without using existing neural network libraries.

\begin{figure}[ht!]
\centering
\includegraphics[width=\linewidth]{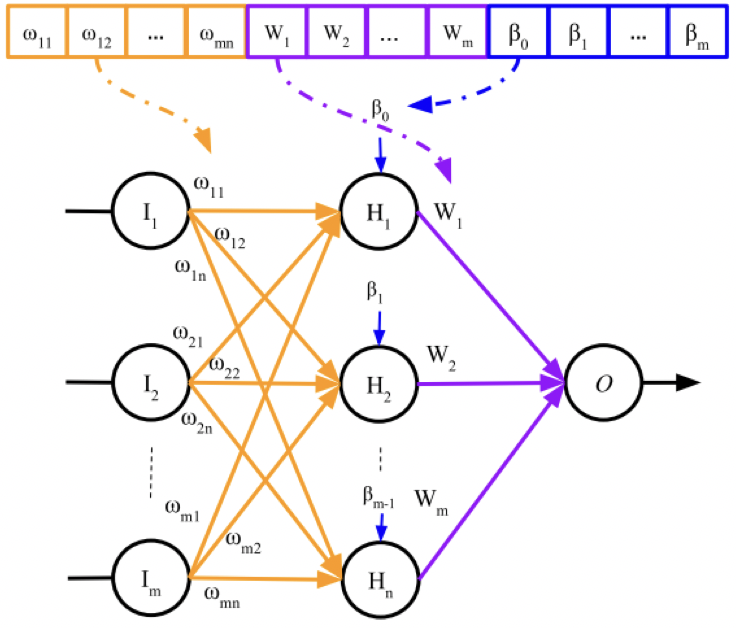}
\caption{Encoding weights as genetic dna encoding vector}
\label{fig:mlp}
\end{figure}

\subsubsection{Evolutionary algorithm}

\begin{figure}[ht!]
\centering
\includegraphics[width=\linewidth]{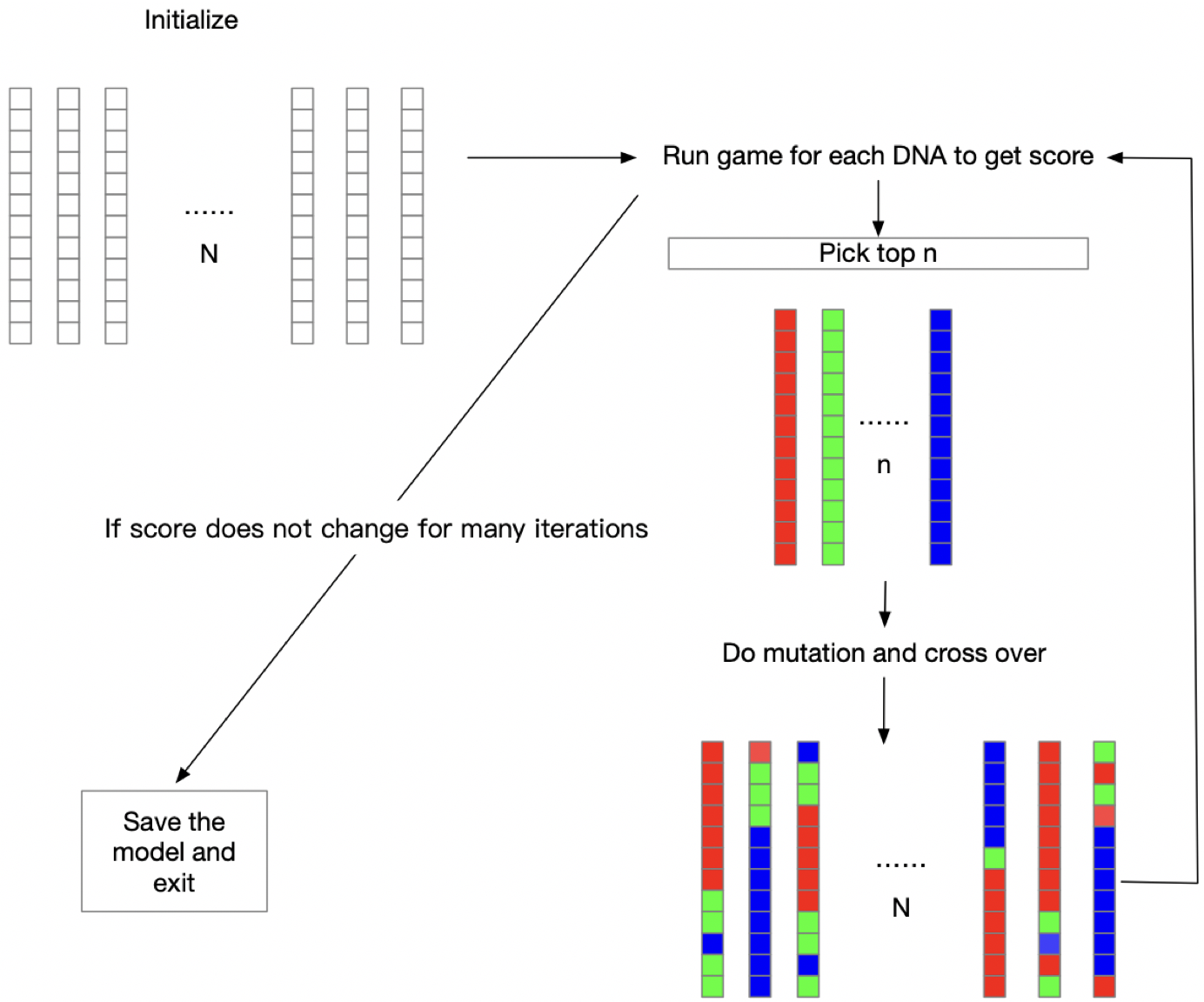}
\caption{Evolutionary Algorithm}
\label{fig:eva}
\end{figure}

For evolutionary algorithm \cite{7955308} , our specific steps are the following: 1. Initialize some number of population of weights using DNA encoding. 2. For each DNA encoding, use neural network to produce the one of the four actions to do and get the final reward from the game until the agent dead and make this reward as fitness value of that DNA encoding in evolutionary algorithm. 3. Select top $n$ DNA encodings of last iteration as parents, use dna mutation method to get the children and calculate the children's fitness value and select top n children as new parents to continue to start a new iteration. Figure \ref{fig:eva} shows the  process. We use multi-processes technique to accelerate the training process. 

\subsubsection{Difference between this approach and cross-entropy method}
 In this approach, we use multi-layer perceptron with non-linear activation function which introduces non-linearity to the model. Besides, this approach can keep the good weights every iteration by crossover which makes the model more stable and consistent. In this algorithm, the weights is not updated by sampling the new distribution, instead the algorithm uses mutation approach to update the weights and the mutation value is randomly generated by random normal distribution with some fixed boundary, so this algorithm has more broad search space. 

\subsection{Double Deep Q-learning Network}
We implement this approach based on the part of code on the Github such as sum tree. We implement the neural network model(based on Keras) and value updating of the double Q-learning on the car racing problem.

\subsubsection{Model input and output}

The input of the model is 96 * 96 * 3 image from the game. The output of the model is: Steering: [-1, 1] which indicates left or right, Gas:  [0, 1]  which indicates if accelerate
and Brake: [0, 1] which indicates if brake. 

\subsubsection{The Principle of DDQN}

\begin{figure}[ht!]
\centering
\includegraphics[width=\linewidth]{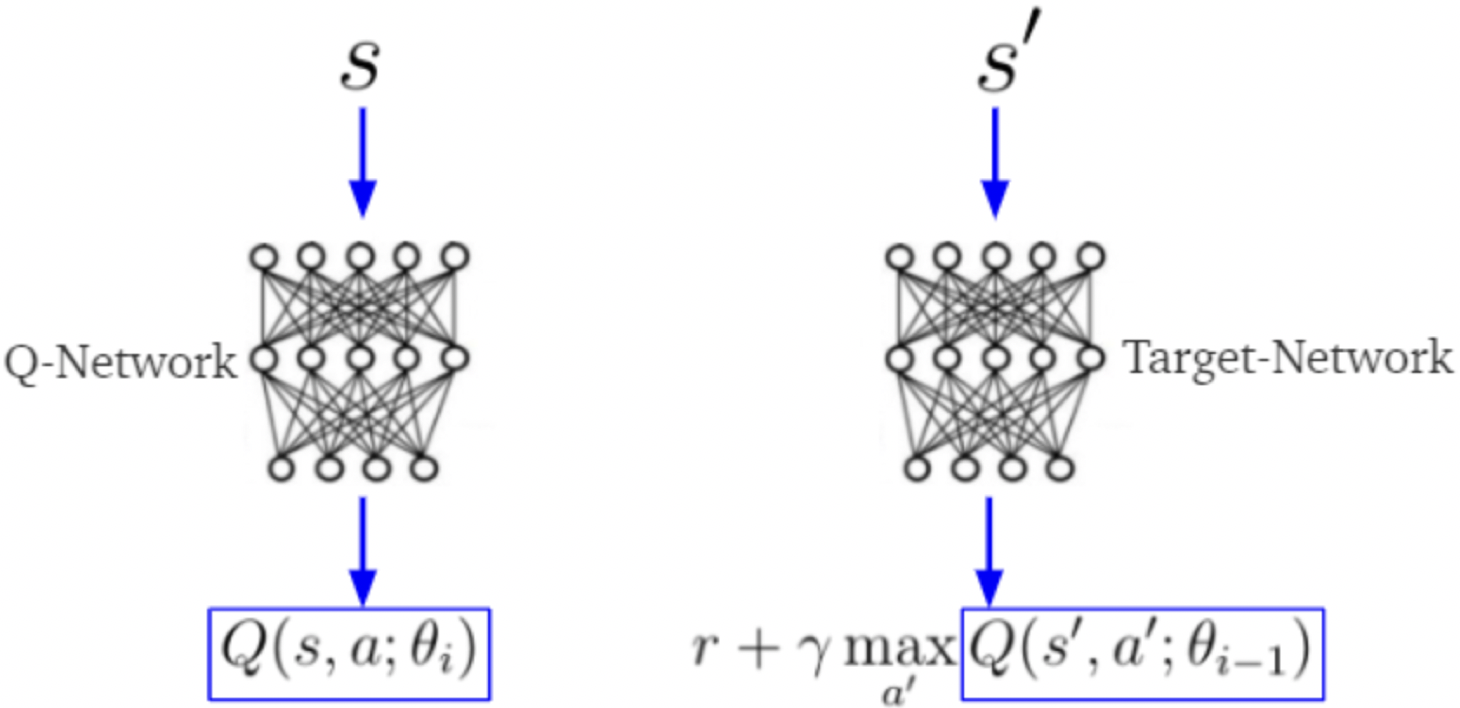}
\caption{Q-network and T-network}
\label{fig:qt}
\end{figure}

Update weights based on two kinds of values, one is Q-value, The Q-value determines choosing which action and the other is T-value, the T-value evaluates the selection. The are two shared weights neural network, one is to compute Q-value and the other is to compute T-value. Figure \ref{fig:qt} shows the Q-network and T-network.

\subsubsection{Network architecture}

\begin{figure}[ht!]
\centering
\includegraphics[width=\linewidth]{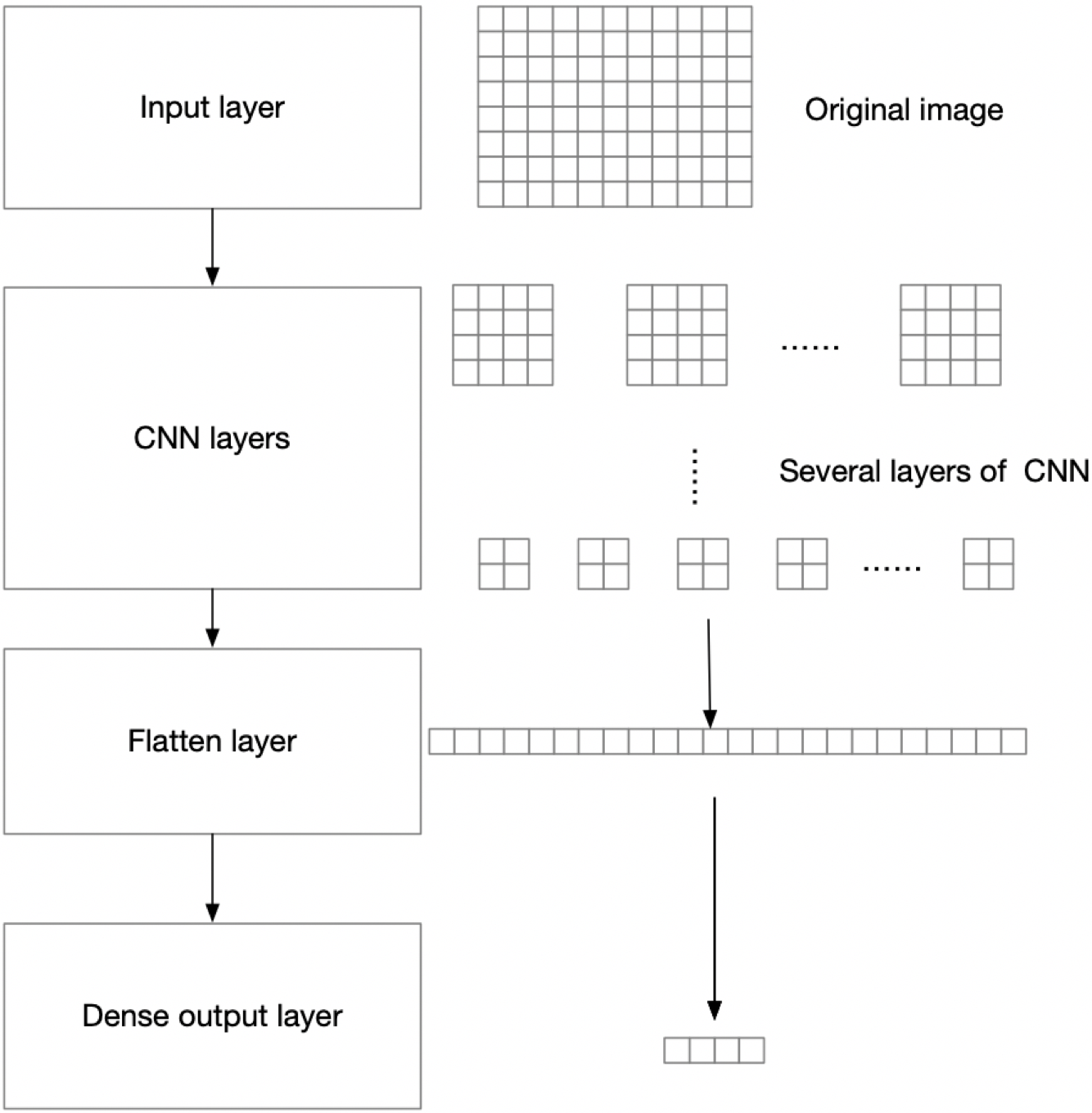}
\caption{Network Architecture}
\label{fig:na}
\end{figure}

This model use original image as input, so we use CNN \cite{LeCun:1999} to extract image features. Figure \ref{fig:na} shows the network architecture of the model.

\section{Results}
\begin{figure}[ht!]
\centering
\includegraphics[width=\linewidth]{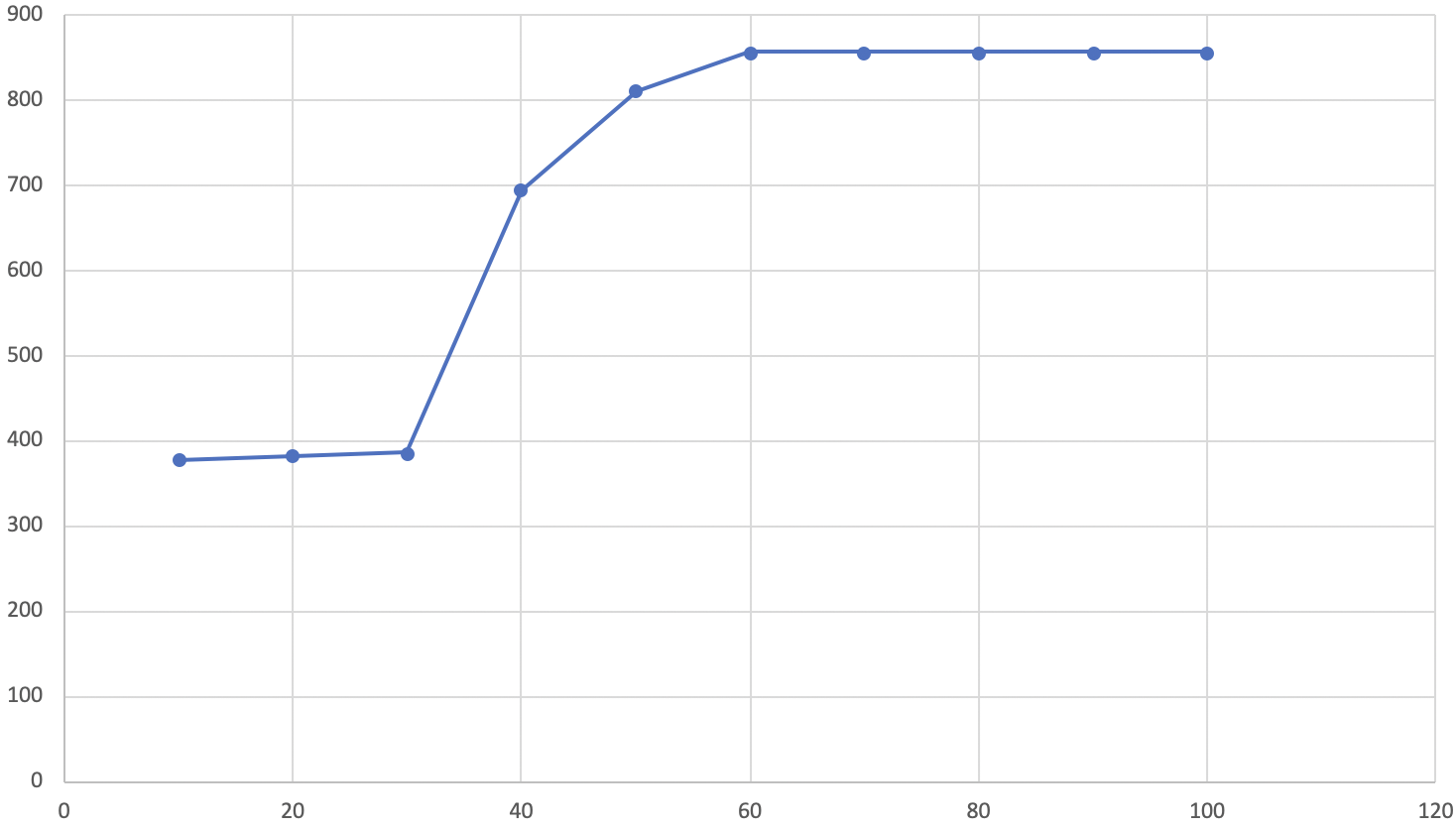}
\caption{Learning curve for genetic MLP}
\label{fig:lce}
\end{figure}

\begin{figure}[ht!]
\centering
\includegraphics[width=\linewidth]{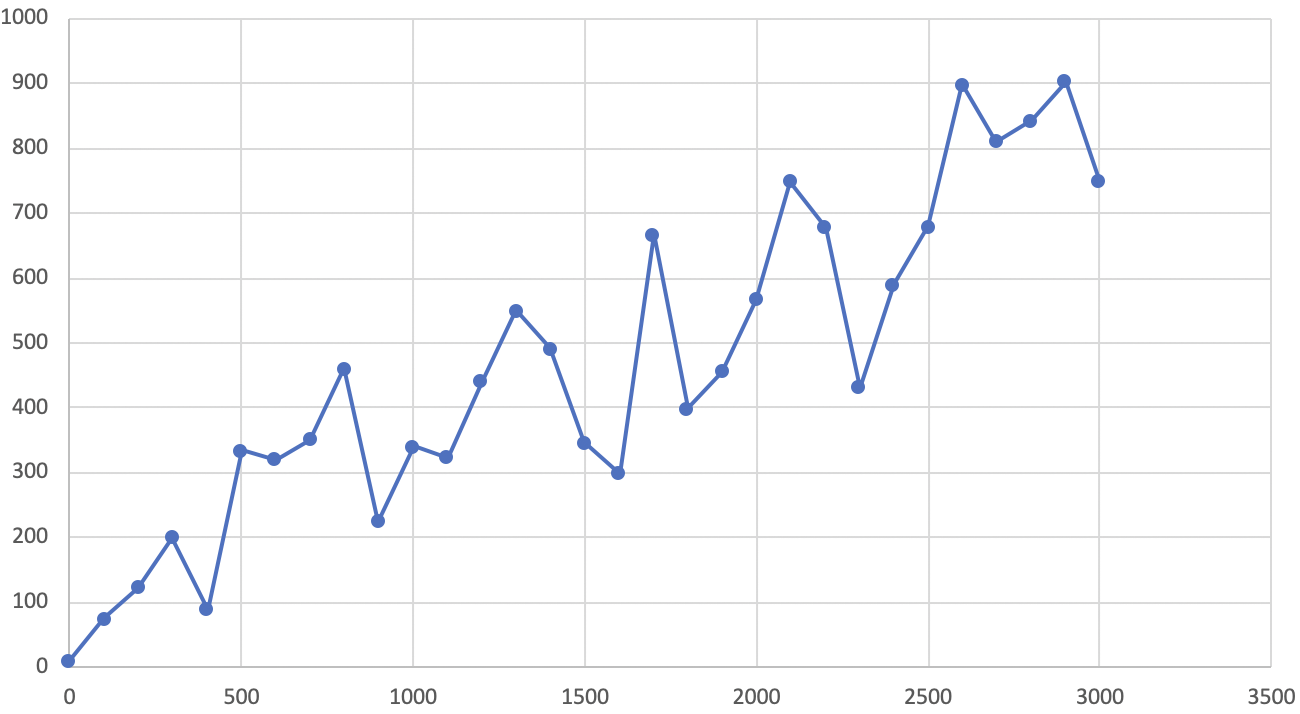}
\caption{DDQN learning curve}
\label{fig:lcd}
\end{figure}

Figure \ref{fig:lce} shows the learning curve of genetic MLP, and Figure \ref{fig:lcd} shows the learning curve of DDQN. After 100 iterations(1 iteration equals N times of game, where N is the number of DNA encoding.) training, the genetic MLP can arrive scores from 856 to 872, and after 1000 iterations, the genetic MLP can arrive scores from 892 to 906. For DDQN, after about 3000 episodes training(1 episode equals to 1 time of replay) to update Q-value and T-value, the value can go up from 900 to 910.

\section{Analysis}

\subsection{Analysis of Evolutionary Algorithm Based Genetic MLP}

For evolutionary algorithm based genetic MLP, we can see it as a kind of search algorithm whose object is to select best weights from a search space by randomly generating value from all possible possibilities of weights combinations within a boundary. It uses mutations to change the weights to make the weight have the possibility to be changed into a better one, and it uses crossover to share the best weights between the different weights combinations which can also reserve the best weights after several iterations. Since for this approach we use the selected feature as our input not the original image shortcut, the model converges quickly and has predictable actions after training such as running as center as possible. And because every iteration, we select the several best DNA encodings, so the score always goes up and never goes down when training. 

The spirit of this approach is similar to the CEM \cite{Rubinstein:2004:CEM:1014902} , however, the main difference is that how they update weights. CEM updates parameters based on sampling the distribution and this approach update weights by mutation and crossover and keep the best weights combinations. Also the non-linearity of MLP is also important. So we assume that if we use the same input as this approach and use CEM, we may get the lower result but we do not have time to prove it.

\subsection{Analysis of Double Deep Q-learning Network}

Double deep Q-learning network is based on deep Q-learning network with the concept of evaluating the Q-learning network by estimating the target value of the problem using the same neural network which makes the network more stable and easy to converge. However, for every episode, the model always needs to do exploration to explore more possibilities which causes sometimes the model can have low rewards during training but after training more and more times the model can get better and better score. Since the model uses original image shortcut as input, the model needs to use CNN to extract image features, so what the models learn is the relation distribution between the output and extracted imaged features. Since its training time is very long, we do not have enough time to train more episodes, but we assume that if we train even more episodes, it can get even better result, however, it still has the top limitation which is related to the features of the maps such as the number of corners, the width of the roads and etc. Image there is a straight road, the model will continuously output up arrow to accelerate but it has its top limitation since the time must be larger than 0.

\section{Conclusion and Future Studies}
In this report, we use two different approaches to solve the car racing problem. We have some findings after acquiring the results and do some analysis. Genetic mlp converges so fast and the result of the genetic mlp will not change after several iterations. DDQN has huge variance when training and it converges slowly and the DDQN can get better result when training more and more episodes but it still has top limitations. Since the genetic mlp converges so fast but cannot get a great score and Q-learning converges slow but can get better result, we may figure a way to try to combine these two methods to make training faster and make the score better. 

\bibliographystyle{aaai}
\bibliography{main} 
\end{document}